\documentclass[10pt,twocolumn,letterpaper]{article}

\usepackage{cvpr} 
\usepackage{CJK}
\usepackage{multicol}
\usepackage{color}
\usepackage{xcolor}
\usepackage{colortbl}
\usepackage{bm}
\usepackage{extarrows}
\usepackage{float}
\usepackage{bbm}
\usepackage{booktabs}
\usepackage{multirow}
\usepackage{amsmath}
\usepackage{amssymb}
\usepackage{algorithm}
\usepackage{algorithmic}
\usepackage{makecell}
\usepackage{amsmath}
\usepackage{amssymb}
\usepackage{amsthm}
\usepackage[utf8]{inputenc}
\usepackage{amsfonts} % \mathbf (z) 符号
\usepackage{caption} 
\usepackage{newfloat}
\usepackage{listings}

\usepackage{xcolor}
\definecolor{thmcolor}{HTML}{0201F5}

\newtheoremstyle{coloredthm}% name
  {3pt}%      Space above
  {3pt}%      Space below
  {\itshape}%         Body font
  {}%         Indent amount (empty = no indent, \parindent = para indent)
  {\bfseries\color{thmcolor}}% Thm head font (Bold AND Colored)
  {.}%        Punctuation after thm head
  {.5em}%     Space after thm head: " " = normal interword space;
  {}%         Thm head spec (can be left empty, meaning 'normal')

\theoremstyle{coloredthm}

\definecolor{cvprblue}{rgb}{0.21,0.49,0.74}
\definecolor{supplcolor}{HTML}{A32D26}

\definecolor{noise20color}{HTML}{FFF9E6} % 极淡橙
\definecolor{noise50color}{HTML}{EDF7ED} % 极淡绿
\definecolor{noise80color}{HTML}{EBF5FB} % 极淡蓝

\usepackage[pagebackref,breaklinks,colorlinks,citecolor=brown]{hyperref}
\usepackage{thmtools}

\def\paperID{*****} % *** Enter the Paper ID here
\def\confName{CVPR}
\def\confYear{2026}

\def\modelname{\mbox{TempRet} }

\newcommand\blfootnote[1]{%
  \begingroup
  \renewcommand\thefootnote{}\footnote{#1}%
  \addtocounter{footnote}{-1}%
  \endgroup
}
\title{TempRet: Temporal Enhancement and Two-Stage Reranking for CVPR 2026 EPIC-KITCHENS-100 Multi-Instance Retrieval Challenge}

\author{Zixu Li$^{1}$~~~~Yupeng Hu$^{1}$\blfootnote{~corresponding authors}~~~~Zhiwei Chen$^{1}$~~~~Zhiheng Fu$^{1}$~~~~Xiaowei Zhu$^{1}$~~~~Weili Guan$^{2}$~~~~Liqiang Nie$^{2}$ \vspace{2mm}\\
$^1$Shandong University\hspace{1.5cm}$^2$Harbin Institute of Technology (Shenzhen)\hspace{1.5cm}\\
{\tt\small \{lizixu.cs, zivczw, fuzhiheng8, xiaoweizhu2005, honeyguan, nieliqiang\}@gmail.com;} \\ 
{\tt\small \ 
huyupeng@sdu.edu.cn}
}
\begin{document}
% \begin{CJK}{UTF8}{gbsn}

\newcommand{\method}{TempRet}
\newcommand{\dataset}{EPIC-KITCHENS-100}
\newcommand{\qwen}{Qwen3-VL-8B}

\maketitle
\begin{abstract}
Video-text retrieval has witnessed remarkable progress driven by large-scale vision-language pretraining, yet most existing approaches inherit an implicit assumption from image-text retrieval: that visual semantics can be captured frame-by-frame.
This assumption overlooks the temporal dynamics of egocentric videos.
The EPIC-KITCHENS-100 Multi-Instance Retrieval (MIR) challenge further raises the bar by providing soft-label relevance matrices rather than binary labels, demanding models that can resolve graded semantic correspondences across modalities.
In this report, we present our solution, termed \textbf{TempRet}, to the CVPR 2026 EPIC-KITCHENS-100 MIR challenge.
Our approach builds upon a CLIP-based dual-encoder backbone and introduces two key components to address the temporal and cross-modal challenges.
First, a \textbf{temporal transformer} operates exclusively on the video side, modeling inter-frame dependencies through learnable positional encodings and multi-head self-attention over frame-level CLIP features.
Second, a \textbf{two-stage reranking} pipeline first retrieves Top-$K$ candidates via the dual-encoder, then refines their scores using a cross-encoder equipped with an Image-Text Matching (ITM) head.
The entire system is trained with Symmetric Multi-Similarity Loss to exploit the soft-label relevance matrices provided by the challenge.
Our method achieves 67.97\% average mAP and 82.92\% average nDCG on the EK-100 MIR benchmark, demonstrating the effectiveness of temporal modeling and cross-modal refinement for egocentric video retrieval.
\end{abstract}

% Video-text retrieval has witnessed remarkable progress driven by large-scale vision-language pretraining, yet most existing approaches inherit an implicit assumption from image-text retrieval: that visual semantics can be captured frame-by-frame. This assumption overlooks the temporal dynamics of egocentric videos. The EPIC-KITCHENS-100 Multi-Instance Retrieval (MIR) challenge further raises the bar by providing soft-label relevance matrices rather than binary labels, demanding models that can resolve graded semantic correspondences across modalities. In this report, we present our solution, termed TempRet, to the CVPR 2026 EPIC-KITCHENS-100 MIR challenge. Our approach builds upon a CLIP-based dual-encoder backbone and introduces two key components to address the temporal and cross-modal challenges. First, a temporal transformer operates exclusively on the video side, modeling inter-frame dependencies through learnable positional encodings and multi-head self-attention over frame-level CLIP features. Second, a two-stage reranking pipeline first retrieves Top-K candidates via the dual-encoder, then refines their scores using a cross-encoder equipped with an Image-Text Matching (ITM) head. The entire system is trained with Symmetric Multi-Similarity Loss to exploit the soft-label relevance matrices provided by the challenge. Our method achieves 67.97% average mAP and 82.92% average nDCG on the EK-100 MIR benchmark, demonstrating the effectiveness of temporal modeling and cross-modal refinement for egocentric video retrieval.    
\section{Introduction}
\label{sec:intro}

Video-text retrieval aims to identify the most relevant video clip from a large corpus given a natural language query.
Unlike image-text retrieval such as image retrieval~\cite{ENCODER,OFFSET,gordo2016deep,HINT,chen2020uniter,MELT,MEDIAN}, video retrieval must model \emph{time}: actions unfold across frames~\cite{ventura2024covr,HUD,gabeur2020multi,REFINE,thawakar2026covr,ReTrack}, objects are manipulated in sequence, and successive moments often determine the meaning of a clip.
This temporal structure is especially important in egocentric videos, where fine-grained hand-object interactions dominate the visual signal.

The EPIC-KITCHENS-100 dataset~\cite{damen2022rescaling} presents a particularly demanding testbed for video-text retrieval.
Its unscripted first-person kitchen videos contain large variation in action speed, camera motion, and object appearance.
The Multi-Instance Retrieval (MIR) challenge further provides a \emph{soft-label relevance matrix} instead of hard binary labels~\cite{STABLE,ERASE}, requiring systems to rank graded semantic correspondences between videos and captions.
This setting reflects the practical ambiguity of egocentric retrieval: several clips may share the same objects, scene context, or action verb, but differ in how well they match a query.

Large-scale vision-language models such as CLIP~\cite{radford2021learning} have demonstrated impressive zero-shot transfer capabilities across a wide range of visual tasks~\cite{PAIR,wang2024sam,HABIT,li2023clip,INTENT}.
However, CLIP is image-centric: when applied frame by frame, it captures strong object and scene semantics but discards ordering, action progression, and causal links across frames.
For MIR, this is a critical limitation because clips may share similar objects and backgrounds while differing in the temporal order or intent of the action~\cite{FineCIR,zhao2017temporal,TEMA,Air-Know}.
One option is to modify the visual backbone with joint spatial-temporal attention, but this increases computation and may disturb the pretrained spatial representations that make CLIP effective.
Another option is simple temporal pooling over frame features, which preserves the backbone but treats the video as an unordered set of observations.
The MIR benchmark calls for a middle ground: keeping the pretrained image-text alignment while adding enough temporal reasoning to distinguish action-centric clips.

In this report, we present \textbf{TempRet}, a method for the CVPR 2026 EK-100 MIR Challenge.
TempRet keeps the CLIP visual and text encoders as efficient pretrained feature extractors, adds a lightweight temporal transformer over frame-level CLS embeddings, and trains the shared retrieval space with Symmetric Multi-Similarity Loss (SMS Loss)~\cite{wang2024sms} to exploit the challenge's soft relevance supervision.
At inference time, TempRet first performs scalable dual-encoder retrieval and then applies a cross-encoder ITM reranker to refine the local ordering of Top-$K$ candidates.
This design separates the roles of the system clearly: the dual encoder provides efficient global retrieval, the temporal transformer improves the video representation before matching, SMS Loss aligns optimization with graded MIR labels, and the reranker performs fine-grained verification only on plausible candidates~\cite{shao2023global,ConeSep}.

In summary, our contributions are as follows:
\begin{itemize}[nosep]
    \item We propose \method, a CLIP-based MIR system that adds temporal reasoning while preserving pretrained spatial-language representations.
    \item We introduce a temporal transformer over frame-level features and a two-stage reranking pipeline for fine-grained candidate discrimination.
    \item We train \modelname with SMS Loss to exploit soft-label relevance matrices, achieving strong performance on the EK-100 MIR benchmark.
\end{itemize}

\section{Method}
\label{sec:method}

\noindent We describe \modelname as a coarse-to-fine video-text retrieval pipeline. The model first builds efficient video and text embeddings with a CLIP-style dual encoder, enriches the video embedding with a lightweight temporal transformer, trains the retrieval space with soft-relevance-aware supervision, and finally applies candidate-level reranking at inference time. The design uses temporal reasoning on the video side, a loss that respects graded relevance, and fine-grained verification over the most likely candidates.

\begin{figure*}[t]
\centering
\includegraphics[width=\textwidth]{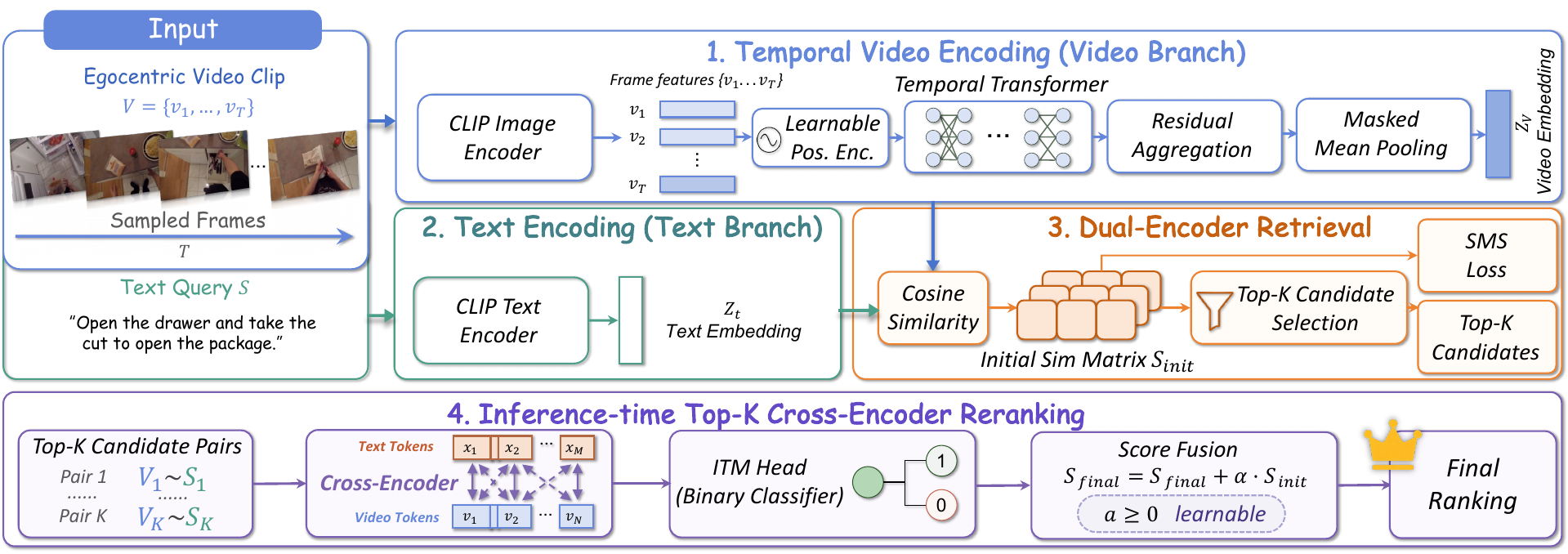}
\caption{Pipeline of \method. In the visual branch, sampled video frames are encoded independently by CLIP image encoder, and passed through a temporal transformer that models inter-frame dependencies. In the text branch, the sentence is encoded by CLIP text encoder. Initial retrieval is performed by cosine similarity between video and text embeddings. The lower part shows the inference-time reranking stage: \modelname selects Top-$K$ candidates from the initial score matrix, applies a cross-encoder with an ITM head to re-score candidate pairs, and fuses the normalized ITM score with the original dual-encoder score to obtain the final ranking.}
\label{fig:pipeline}
\end{figure*}

% ------------------------------------------------
\subsection{Visual Encoder}
\label{sec:visual}

We adopt CLIP's image encoder as our visual backbone. For a video of $T$ frames, each frame is independently encoded. This choice deliberately preserves the strong spatial and object-level semantics learned by CLIP, which are valuable for kitchen scenes containing recurring tools, ingredients, containers, and hand-object configurations. From each frame, we extract the CLS token output and apply the visual projection matrix to map it to the shared embedding dimension $D = 512$:
\begin{equation}
    \mathbf{v}_t = \text{ImageEncoder}(V_t) \cdot \mathbf{W}_{\text{proj}}, \quad t = 1, \dots, T.
\end{equation}
This produces a sequence of frame-level embeddings $\mathbf{X} = [\mathbf{v}_1, \dots, \mathbf{v}_T] \in \mathbb{R}^{T \times D}$, which serves as input to the temporal transformer. At this point, each feature is still frame-local: it tells the model what is visible in a frame, but not how the observed state relates to earlier or later frames. The next module is therefore responsible for converting these independent observations into a temporally aware clip representation.

% ------------------------------------------------
\subsection{Temporal Transformer}
\label{sec:temporal}

The original CLIP ViT processes each image independently, and it has no mechanism for modeling temporal relationships between frames. This limitation is significant for egocentric retrieval because many queries describe actions rather than static scenes. A clip showing a hand reaching toward a cup, grasping it, and placing it elsewhere may contain nearly the same objects as another clip, but the retrieval target depends on the temporal progression of the interaction. To address this, we introduce a dedicated \textbf{temporal transformer} that operates on the video side, modeling inter-frame dependencies before the video representation is compared with text.

Before entering the transformer, the frame sequence is augmented with learnable positional encodings. These encodings make temporal order explicit, preventing self-attention from treating the frame sequence as an unordered set. The module therefore retains the frame-level semantics produced by CLIP while adding the missing information about when each visual observation occurs within the sampled clip.

\paragraph{Multi-Head Self-Attention.} The sequence $\mathbf{X}$ is processed by a stack of $L$ transformer layers.
Each frame attends to all other frames in the clip, enabling the model to capture temporal dependencies such as action onset, object manipulation, and hand-object interaction patterns:
\begin{equation}
    \mathbf{H}^{(l)} = \mathbf{H}^{(l-1)} + \text{MHA}\big(\text{LN}(\mathbf{H}^{(l-1)})\big).
\end{equation}

\paragraph{Feed-Forward Process.}
Each attention output passes through a position-wise MLP with 4$\times$ expansion and GELU activation. This feed-forward block increases the capacity of each temporal layer after the attention step has exchanged information across frames:
\begin{equation}
    \mathbf{H}^{(l)} = \mathbf{H}^{(l)} + \text{MLP}\big(\text{LN}(\mathbf{H}^{(l)})\big)
\end{equation}
where $\text{MLP}(\mathbf{x}) = \mathbf{W}_2 \cdot \text{GELU}(\mathbf{W}_1 \mathbf{x})$, with $\mathbf{W}_1 \in \mathbb{R}^{4D \times D}$ and $\mathbf{W}_2 \in \mathbb{R}^{D \times 4D}$.

\paragraph{Why a Separate Temporal Transformer?}
An alternative design would be to extend the ViT itself with temporal attention (e.g., joint spatial-temporal attention). However, we choose a separate lightweight transformer on top of per-frame CLS tokens for three reasons. It preserves the pretrained ViT's spatial representations and it is parameter-efficient: temporal reasoning is learned over a compact sequence of frame embeddings rather than over all patch tokens. Third, it allows flexible clip lengths at inference without retraining the visual backbone, since the temporal module operates after frame encoding.

\paragraph{Residual Connection and Temporal Aggregation}

The transformer output is combined with the original frame features via a residual connection, preserving the pretrained frame-level semantics:
\begin{equation}
    \tilde{\mathbf{X}} = \text{Transformer}(\mathbf{X}) + \mathbf{X}.
\end{equation}
This residual path is important because the temporal transformer is meant to complement, not replace, CLIP's frame features. The video-level representation is then obtained via masked mean pooling over the temporal dimension:
\begin{equation}
    \tilde{\mathbf{v}} = \frac{\sum_{t=1}^{T} m_t \cdot \tilde{\mathbf{x}}_t}{\sum_{t=1}^{T} m_t}, \quad \tilde{\mathbf{v}} \in \mathbb{R}^D.
\end{equation}
where $m_t \in \{0, 1\}$ is the video mask indicating whether frame $t$ is valid.

% ------------------------------------------------
\subsection{Text Encoder}

We use CLIP's text encoder to encode the input caption. The text encoder produces a sequence of token embeddings, from which we extract the EOS token embedding as the sentence-level representation. This representation summarizes the query in the same embedding space used by the visual branch. The text embedding is projected to the shared dimension $D$ via the text projection matrix:
\begin{equation}
    \mathbf{s} = \text{TextEncoder}(S) \cdot \mathbf{W}_{\text{txt\_proj}}, \quad \mathbf{s} \in \mathbb{R}^D.
\end{equation}

\subsection{Training with Symmetric Multi-Similarity Loss}
\label{sec:loss}

To exploit the soft-label relevance matrix provided by the MIR challenge, we follow the Symmetric Multi-Similarity Loss (SMS Loss)~\cite{wang2024sms}. This loss is well matched to the benchmark because MIR relevance is graded: a candidate may be semantically close to the query even when it is not the exact paired item. Given a batch of $N$ video-text pairs, the model computes a retrieval similarity matrix $\mathbf{S} \in \mathbb{R}^{N \times N}$ from the embeddings.

SMS Loss extends the standard max-margin ranking loss by incorporating sample-pair weights derived from the relevance matrix $\mathbf{R}$. For each pair $(i, j)$, the loss function distinguishes three cases based on the weight $w_{ij}$:
\begin{equation}
    \mathcal{L}_{ij} = \begin{cases}
        \text{ReLU}(w_{ij} \cdot m - \Delta_{ij}) & \text{if } w_{ij} > \epsilon, \\
        \text{ReLU}(|w_{ij} \cdot m - \Delta_{ij}| - \tau) & \text{if } |w_{ij}| \leq \epsilon, \\
        \text{ReLU}(-(w_{ij} \cdot m - \Delta_{ij})) & \text{if } w_{ij} < -\epsilon,
    \end{cases}
\end{equation}
where $\Delta_{ij} = s_{ii} - s_{ij}$ is the margin between the positive pair score and the negative pair score, $m = 0.2$ is the margin hyperparameter, and $\tau = 0.1$ is the threshold for neutral pairs.

The three cases prevent the training objective from over-simplifying MIR supervision. Highly relevant pairs are encouraged to stay close, clearly irrelevant pairs are pushed away, and neutral or weakly related pairs are handled more cautiously. This is important for EPIC-KITCHENS retrieval because many clips share the same kitchen environment and object vocabulary, while differing only in action intent or temporal order.

% ------------------------------------------------
\subsection{Two-Stage Reranking}
\label{sec:rerank}

The dual-encoder computes initial video-text similarity scores efficiently, but treats each modality independently. To capture fine-grained cross-modal interactions that the dual-encoder misses, we adopt a two-stage reranking strategy: coarse retrieval with the dual-encoder followed by fine-grained re-scoring with a cross-encoder. The motivation is practical: exhaustive cross-encoder scoring over all video-query pairs is too expensive, while applying it only to the most plausible candidates provides a strong accuracy-efficiency trade-off.

\paragraph{Stage 1: Coarse Retrieval.}
Given all video and text embeddings, we compute the initial similarity matrix $\mathbf{S}_{\text{init}} \in \mathbb{R}^{N_q \times N_v}$ via cosine similarity. For each query, we select the Top-$K$ candidate videos by score. This stage relies on the temporally enhanced video representations described above, so the candidate set already reflects video-side temporal structure before the more expensive pairwise reranker is applied.

\paragraph{Stage 2: Cross-Encoder Re-scoring.}
For each query $q_i$ and its Top-$K$ candidates $\{c_{j_1}, \dots, c_{j_K}\}$, we concatenate their embeddings and pass each pair through a cross-encoder that models token-level interactions between the video and text:
\begin{equation}
    \mathbf{o}_{ij} = \text{CrossEncoder}([\mathbf{v}_j ; \mathbf{q}_i]), \quad \mathbf{o}_{ij} \in \mathbb{R}^D.
\end{equation}

The cross-encoder output is fed to a binary ITM classification head that outputs a matching probability:
\begin{equation}
    s_{ij}^{\text{ITM}} = \text{softmax}\big(\text{ITM}(\mathbf{o}_{ij})\big)_1.
\end{equation}
Only the Top-$K$ positions in the score matrix receive ITM scores; all other positions are initialized to a sentinel value.

\paragraph{Miss Penalty and Normalization.}
For positions not in the Top-$K$, we replace the sentinel with the per-row minimum ITM score, ensuring non-candidates receive the lowest score from that query's perspective:
\begin{equation}
    s_{ij}^{\text{ITM}} \leftarrow \min_k s_{ik}^{\text{ITM}}, \quad \text{if } j \notin \text{Top-}K(q_i),
\end{equation}
This miss penalty keeps the score matrix complete while preserving the reranker's intended scope: candidates outside the retrieved set should not receive artificial advantages from missing ITM values.

\paragraph{Score Fusion.}
The final score combines the ITM score with a residual of the initial similarity:
\begin{equation}
    \mathbf{S}_{\text{final}} = \mathbf{S}_{\text{ITM}}^{\text{norm}} + \alpha \cdot \mathbf{S}_{\text{init}}^{\text{norm}}.
\end{equation}
The residual term keeps the broad semantic signal from the dual encoder, while the normalized ITM score contributes more detailed pairwise evidence. In this way, the final ranking remains anchored to the efficient retrieval space but can still correct local ordering errors among visually or semantically similar candidates.

\section{Experiments}
\label{sec:experiments}

\subsection{Dataset and Setup}

\begin{table*}[t]
\centering
\caption{Official leaderboard results on the EPIC-KITCHENS-100 MIR benchmark. The last row is \method.}
\label{tab:leaderboard_results}
\begin{tabular}{@{}lcccccc@{}}
\toprule
\textbf{Method} & \textbf{mAP-Avg} & \textbf{mAP-T2V} & \textbf{mAP-V2T} & \textbf{nDCG-Avg} & \textbf{nDCG-T2V} & \textbf{nDCG-V2T} \\
\midrule
Privacy & 66.21 & 60.71 & 71.70 & 78.59 & 76.75 & 80.43 \\
Enid & 66.78 & 61.75 & 71.81 & 82.08 & 80.25 & 83.91 \\
\textbf{TempRet} & \textbf{67.97} & \textbf{65.03} & \textbf{70.91} & \textbf{82.92} & \textbf{81.41} & \textbf{84.42} \\
\bottomrule
\end{tabular}
\end{table*}

We evaluate on the EPIC-KITCHENS-100 MIR benchmark, which measures retrieval quality in both Video-to-Text (V2T) and Text-to-Video (T2V) directions. The benchmark is particularly suitable for testing our design because it contains egocentric kitchen videos with dense object interactions and soft relevance annotations. Good performance therefore requires both accurate temporal video representations and a ranking objective that can exploit graded similarity.

We use ViT-L/14 as the visual backbone with input resolution 224$\times$224. In the reranking stage, $K$ is set to $1000$ and $\alpha$ is $0.002$. The model is fine-tuned on the training dataset. Training uses AdamW optimizer with learning rate $1.8 \times 10^{-5}$, batch size 64, and cosine LR scheduling. The temporal transformer uses $L=4$ layers with $D=512$ and 8 heads, receiving 2$\times$ learning rate. This differential learning rate allows the newly introduced temporal layers to adapt quickly while keeping the pretrained visual and text encoders stable.

\subsection{Evaluation Metrics}
The MIR Challenge reports two ranking metrics: mean Average Precision (mAP) and normalized Discounted Cumulative Gain (nDCG). mAP evaluates whether relevant items are ranked early across the retrieved list, making it sensitive to the precision of high-confidence matches. nDCG further accounts for graded relevance by discounting lower-ranked items and rewarding systems that place highly relevant videos or captions near the top. Since the benchmark provides soft relevance matrices, nDCG is especially important for measuring whether the predicted ranking respects different degrees of semantic match rather than only separating positives from negatives.

\subsection{Results}

Table~\ref{tab:leaderboard_results} summarizes the official leaderboard results. The last row corresponds to \method. Among the listed teams, \modelname achieves the best average performance on both summary metrics, with 67.97\% mAP-Avg and 82.92\% nDCG-Avg. The gains are strongest in the Text-to-Video direction, where \modelname reaches 65.03\% mAP and 81.41\% nDCG. In the Video-to-Text direction, \modelname obtains 70.91\% mAP and the best listed nDCG of 84.42\%, indicating that the final ranking better respects the benchmark's graded relevance annotations.

\subsection{Ablation Study}

Table~\ref{tab:ablation} reports our internal ablation study for the main components shown in Fig.~\ref{fig:pipeline}. Starting from the full \modelname system, we remove the reranking stage, the temporal transformer, or both components while keeping the same evaluation protocol. This table therefore directly corresponds to the method diagram: the first ablation tests the lower reranking block, the second tests the video-side temporal transformer, and the final setting removes both temporal modeling and reranking. The results show that temporal modeling is the primary source of improvement, while reranking provides a smaller but consistent gain when the temporal dual-encoder representation is available.

\begin{table}[t]
\centering
\caption{Ablation study on the EPIC-KITCHENS-100 MIR benchmark. We report average mAP and average nDCG over T2V and V2T.}
\label{tab:ablation}
\begin{tabular}{@{}lcc@{}}
\toprule
\textbf{Setting} & \textbf{Avg mAP} & \textbf{Avg nDCG} \\
\midrule
\textbf{TempRet} & \textbf{67.97} & \textbf{82.92} \\
w/o Reranking & 66.21 & 81.10 \\
w/o Temporal Transformer & 55.61 & 68.42 \\
w/o Temporal\&Reranking & 56.61 & 67.40 \\
\bottomrule
\end{tabular}
\end{table}
\subsection{Analysis}

\paragraph{Effect of Temporal Modeling.} The temporal transformer is the most important component in \method. Removing it decreases the average mAP from 67.97\% to 55.61\% and the average nDCG from 82.92\% to 68.42\%, corresponding to drops of 12.36 and 14.50 points, respectively. Removing both temporal modeling and reranking gives a similarly low result, 56.61\% average mAP and 67.40\% average nDCG, confirming that frame-level CLIP features alone are insufficient for the MIR setting. Without explicit temporal information, the model mainly relies on static frame semantics and cannot reliably distinguish action progression, manipulation order, or short temporal transitions.

\paragraph{Effect of Reranking.} Removing the reranking stage reduces the average mAP from 67.97\% to 66.21\% and the average nDCG from 82.92\% to 81.10\%. The drop is smaller than that caused by removing the temporal transformer, which is expected because reranking operates only after the dual encoder has already produced a Top-$K$ candidate set. Nevertheless, the consistent decrease shows that candidate-level re-scoring improves fine-grained discrimination, especially for nDCG, where placing highly relevant candidates at the top of the ranked list matters more than retrieving broadly related clips.

\section{Conclusion}
\label{sec:conclusion}

We presented \modelname for the EPIC-KITCHENS-100 MIR Challenge, combining a CLIP-style dual encoder, a lightweight temporal transformer, SMS Loss for soft relevance supervision, and Top-$K$ cross-encoder reranking. The results show that temporal modeling is crucial for egocentric video retrieval, while reranking further improves the local ordering of difficult candidates. Together, these components provide a practical coarse-to-fine retrieval pipeline for action-centric videos with graded relevance annotations.

{
    \small
    \bibliographystyle{ieeenat_fullname}
    \bibliography{main}
}

% WARNING: do not forget to delete the supplementary pages from your submission 
% \end{CJK}
\end{document}